\pdfoutput=1
\documentclass[11pt]{article}

\usepackage[final]{acl}

\usepackage{times}
\usepackage{latexsym}

\usepackage[T1]{fontenc}
\usepackage[utf8]{inputenc}

\usepackage{microtype}

\usepackage{inconsolata}

\usepackage{graphicx}

\usepackage{url}
\usepackage{subfigure}
\usepackage{multirow}
\usepackage{enumitem}
\usepackage{stmaryrd}
\usepackage{array}
\usepackage{threeparttable}
\usepackage{blindtext}
\usepackage{amssymb}
\usepackage{fontawesome}

\usepackage[ruled,vlined,linesnumbered,longend]{algorithm2e}
\usepackage{svg}
\usepackage{placeins}

\usepackage{soul}
\usepackage[utf8]{inputenc}
\usepackage{graphicx}
\usepackage{amsmath}
\usepackage{booktabs}
\usepackage{lipsum}
\usepackage{listings}
\usepackage{xcolor, soul}
\sethlcolor{blue!10}
\usepackage{colortbl}
\usepackage{bbm}

\usepackage{epigraph}

\usepackage{cleveref}
\crefformat{section}{\S#2#1#3} % see manual of cleveref, section 8.2.1
\crefformat{subsection}{\S#2#1#3}
\crefformat{subsubsection}{\S#2#1#3}

\usepackage{hyperref}
\usepackage{CJKutf8}
\newcommand{\method}{\textsc{LinguaLens}\xspace}
\newcommand{\data}{\textsc{LinguaLens-Data}\xspace}

\definecolor{sgreen}{HTML}{F3FADF}  % lightest
\definecolor{mgreen}{HTML}{E0EAB5}  % medium-light
\definecolor{dgreen}{HTML}{CDDC8C}  % medium-dark
\definecolor{ddgreen}{HTML}{B8CF61} % darkest

\definecolor{mymelon}{RGB}{254,136,99}

\usepackage{amsmath,amsfonts,bm}

\def\eqref#1{equation~\ref{#1}}
\def\1{\bm{1}}

\DeclareMathAlphabet{\mathsfit}{\encodingdefault}{\sfdefault}{m}{sl}
\SetMathAlphabet{\mathsfit}{bold}{\encodingdefault}{\sfdefault}{bx}{n}

\title{
LinguaLens: Towards Interpreting Linguistic Mechanisms \\
of Large Language Models via Sparse Auto-Encoder
}

\author{
  Yi Jing$^\spadesuit$, Zijun Yao$^\spadesuit$, Hongzhu Guo$^\heartsuit$, Lingxu Ran$^\spadesuit$, Xiaozhi Wang$^\spadesuit$, Lei Hou$^\spadesuit$, Juanzi Li$^\spadesuit$\footnotemark[4] \\
  $^\spadesuit$DCST, BNRist; KIRC, Institute for Artificial Intelligence, Tsinghua University, \textit{China} \\
  $^\heartsuit$Department of Chinese Language and Literature, Peking University, \textit{China} \\
  \texttt{jingy22@mails.tsinghua.edu.cn, lijuanzi@tsinghua.edu.cn}
}
\begin{document}
\maketitle

\renewcommand{\thefootnote}{\fnsymbol{footnote}}
\footnotetext[4]{Corresponding author.}

\begin{abstract}
Large language models (LLMs) demonstrate exceptional performance on tasks requiring complex linguistic abilities, such as reference disambiguation and metaphor recognition/generation. Although LLMs possess impressive capabilities, their internal mechanisms for processing and representing linguistic knowledge remain largely opaque. Prior research on linguistic mechanisms is limited by coarse granularity, limited analysis scale, and narrow focus. In this study, we propose \method, a systematic and comprehensive framework for analyzing the linguistic mechanisms of large language models, based on Sparse Auto-Encoders (SAEs). We extract a broad set of Chinese and English linguistic features across four dimensions—morphology, syntax, semantics, and pragmatics. By employing counterfactual methods, we construct a large‐scale counterfactual dataset of linguistic features for mechanism analysis. Our findings reveal intrinsic representations of linguistic knowledge in LLMs, uncover patterns of cross‐layer and cross‐lingual distribution, and demonstrate the potential to control model outputs. This work provides a systematic suite of resources and methods for studying linguistic mechanisms, offers strong evidence that LLMs possess genuine linguistic knowledge, and lays the foundation for more interpretable and controllable language modeling in future research.
\end{abstract}

\begin{center}
    \renewcommand{\arraystretch}{1.4} 
    \begin{tabular}{c@{\hspace{0.5em}}l@{\hspace{0.7em}}l}
        \faGithub & Code & \href{https://github.com/THU-KEG/LinguaLens}{THU-KEG/LinguaLens} \\
        \faDatabase & Dataset & \href{https://huggingface.co/datasets/THU-KEG/LinguaLens-Data}{THU-KEG/LinguaLens-Data}
    \end{tabular}
\end{center}

\section{Introduction}
Large language models (LLMs) demonstrate strong performance on tasks requiring different levels of linguistic competence, such as dependency parsing~\cite{lin2022dependency,benchclamp}, reference disambiguation~\cite{disambiguation}, and metaphor interpretation~\cite{gpt3-metaphor,eval-non-literal-intent,metaphor-reasoning}.

\begin{figure}[tp]
    \centering
    \includegraphics[width=0.98\linewidth]{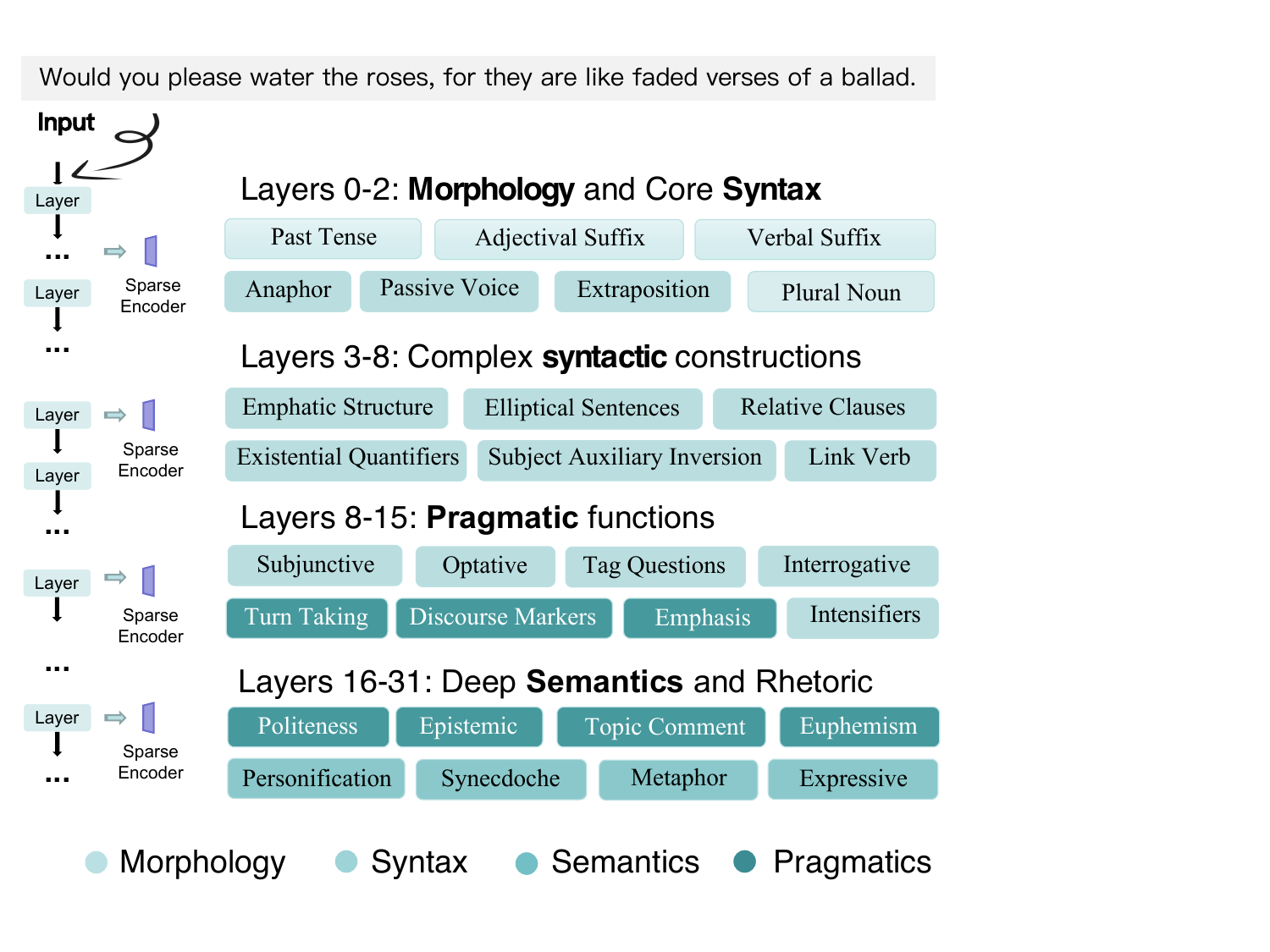}
    \caption{The main linguistic features activated at different layers are observed when example sentences are input to the model. Through a Sparse Auto-Encoder, each layer’s activation values are mapped into a sparse space and the basis vectors corresponding to predefined linguistic features are extracted. According to the results, the model’s 32 layers are divided into four stages, in order: Morphology and Core Syntax, Complex Syntactic Constructions, Pragmatic Functions, and Deep Semantics and Rhetoric.}
    \label{fig:intro}
\end{figure}

Although their linguistic abilities are often attributed to emergent capabilities from large‑scale pretraining and model scale~\cite{manning2020emergent,allen2023physics,mahowald2024dissociating}, the underlying mechanisms by which LLMs process these linguistic structures remain underexplored and lack systematic explanation~\cite{saba2023stochastic}. Therefore, our goal is to interpret the linguistic mechanisms of LLMs by addressing the following questions: \textit{(1) Can we identify the minimal components within an LLM responsible for specific linguistic processing abilities? (2) Can we comprehensively model the internal linguistic functionalities of the model?}

Prior attempts to explain LLM linguistic mechanisms typically rely on expert‑designed prompts that ask the model to elucidate its generation process~\cite{yin2022interpreting}. However, such behavior‑based approaches do not provide structure‑level mechanistic insights. More recent work seeks to link specific linguistic capabilities to internal structures—such as hidden states~\cite{katz2023visit}, attention heads~\cite{wu2020structured}, and activated neurons~\cite{sajjad2022neuron,huang2023rigorously}—but they face two main challenges:

\textbf{Coarse interpretive granularity.}  
Mechanistic interpretation aims to uncover \textit{atomic} linguistic structures within LLMs. Yet even neurons—the finest native components—exhibit poly‑semantic activations, responding to multiple conditions~\cite{yan2024encourage}. This necessitates extracting finer‑grained structures to truly interpret linguistic mechanisms.

\textbf{Limited analysis scale.}  
Existing studies focus on one or a few linguistic features, often within a single subfield (e.g., syntax or semantics), neglecting large‑scale, systematic analysis across diverse linguistic phenomena. A scalable, automated framework is needed to interpret language mechanisms comprehensively.

To address these challenges, we propose \method, a framework that utilizes a sparse auto‑encoder (SAE) to interpret LLM linguistic mechanisms. The SAE learns a projection matrix that decomposes LLM hidden states into an extremely high‑dimensional feature space under a sparsity constraint, where each dimension captures a single semantic concept (Figure~\ref{fig:intro}). \method comprises three modules:  
1. Construction of a large‑scale, multilingual, counterfactual linguistic dataset to support systematic discovery of linguistic structures;  
2. Sparse feature analysis to interpret the SAE‑extracted features, providing fine‑grained and comprehensive mechanistic insights;  
3. Feature intervention, manipulating LLM behavior via targeted interventions on interpretable features to verify causal relationships and enable controlled steering of language behavior.

Specifically, we first build a large-scale hierarchical counterfactual linguistic dataset with annotated corpora, categorizing features into morphology, syntax, semantics, and pragmatics. These widely studied linguistic abilities ensure the feasibility of interpretability. We automate feature extraction via SAE activation analysis and an LLM‑based agent, and introduce a causal analysis method that intervenes on SAE base vectors with an LLM judge to evaluate effects. Building on this, we analyze cross‑layer function distribution and cross‑lingual representation patterns differences of linguistic features.

We conduct extensive experiments on Llama‑3.1‑8B~\cite{grattafiori2024llama3}. Our results demonstrate that \method can effectively identify linguistic competence features at scale, laying the groundwork for further systematic analysis.  

\section{Related Works}

Linguistic mechanism interpretation has been a ever-chasing goal since the emergence of LLMs. Researchers build linguistic datasets to evaluate the linguistic capability and to interpret linguistic mechanisms.
We review linguistic datasets for LLMs and corresponding mechanistic interpretation works.
We will also introduce the basic concepts for sparse auto-encoder.

\textbf{Linguistic Datasets for LLMs.}
Previous studies have introduced numerous linguistic datasets for large‑model research, which can be divided into two main categories. 
The first comprises minimal‑pair challenge sets—such as BLiMP~\cite{warstadt2020blimp}, CLiMP~\cite{xiang2021climp}, and SyntaxGym~\cite{gauthier2020syntaxgym}—that use acceptability judgments to evaluate morphosyntactic competence. 
The second consists of counterfactual or contrastive corpora—including CAD~\cite{sen2022counterfactually}, Contrast Sets~\cite{gardner2020contrastsets}, and Polyjuice~\cite{wu2021polyjuice}—that assess model by generating factual/counterfactual pairs. 
These resources focus primarily on syntactic analysis and performance evaluation, and are not suited for systematic investigation of models’ internal linguistic representations.

\textbf{Linguistic Mechanism Interpretation.}
Previous work has employed a variety of methods to study linguistic mechanisms in large language models, including attention head analysis~\cite{bertlook19},  probing classifiers~\cite{belinkov2022probing,he2024decoding}, causal intervention techniques~\cite{finlayson2021causal,hao2023verb}, and neuron‑level analyses~\cite{sajjad2022neuron}. However, these approaches have not been applied in a unified, large‑scale framework to systematically chart models’ full range of linguistic capabilities.

\textbf{Sparse Auto-encoder.} 
Recent work has employed sparse auto-encoders (SAEs) to interpret the hidden-layer activations of large language models by decomposing them into a large set of concept features~\cite{gao_sparse}. 
These concept features exhibit mono-semanticity and hold considerable interpretability potential~\cite{cunningham2023sparse}.
In particular, an SAE maps the hidden states $\mathbf{f} \in \mathbb{R}^{d}$ in LLMs into the feature space with sparse activations:
\begin{equation*}
    \mathbf{f} = \text{SparseConstraint} \left( \mathbf{W}_e \mathbf{h} + \mathbf{b}_e \right),
\end{equation*}
where the SAE is parameterized by $\mathbf{W}_e \in \mathbb{R}^{(r \times d) \times d}, \mathbf{b}_e \in \mathbb{R}^{(r \times d)}$.
$r$ is the expansion ratio, defined as the factor by which the hidden state dimension is expanded.
Commonly used sparse constraint include TopK~\cite{gao_sparse} and JumpReLU~\cite{rajamanoharan2024improving} functions.
As each dimension of the sparse activation in $\mathbf{f}$ corresponds to a base vector in $\mathbf{W}_e$, this paper uses base vector to denote features extracted by SAE.

% However, interventions based on SAE-extracted features have yielded suboptimal results, and existing studies suggest that these concept features may be distributed across multiple layers.

\section{Methodology}

\method consists of three key components.
(1) A multi‑level counterfactual dataset of linguistic features supporting systematic linguistic mechanism analysis;
(2) An SAE‑based linguistic feature extraction method leveraging LLM agents and correlation analysis;
(3) A Linguistic feature intervention method for causality validation and LLM steering.

\begin{figure*}[tp]
    \centering
    \includegraphics[width=0.97\textwidth]{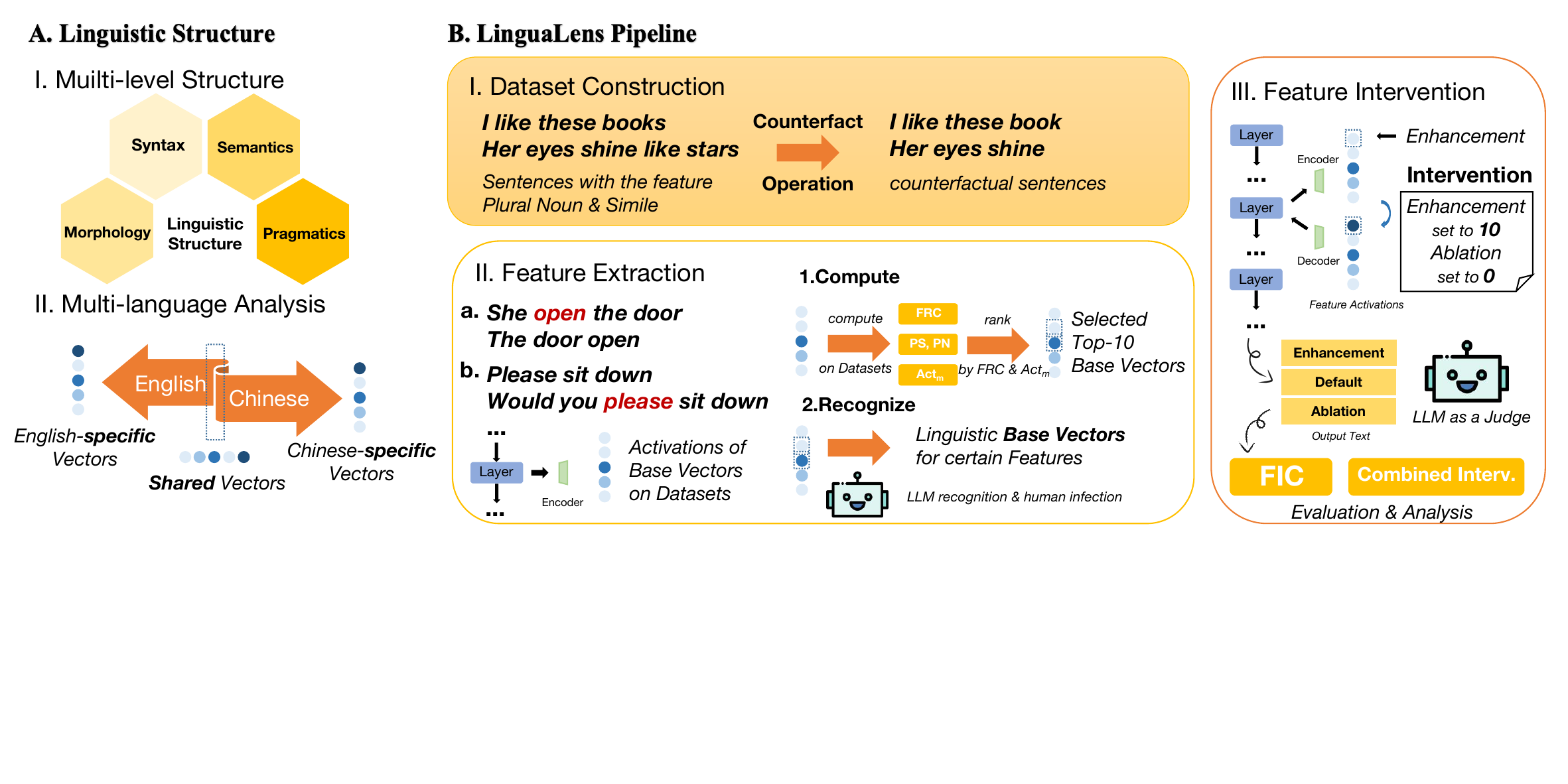}
    \vspace{-0.01in}
    \caption{
    The overall framework of \method.
    We propose a framework for the linguistic mechanisms of large-scale models that encompasses four dimensions of theoretical linguistics and a cross‑lingual analysis of both Chinese and English. 
    The experimental workflow is as follows: 
    (1) Construct counterfactual datasets; 
    (2) Extract features by analyzing the activation values of base vectors on the datasets; 
    (3) Intervene in the model output by modifying activation values and assess causality using an LLM as a judge.
    }\label{fig:method}
\end{figure*}

\subsection{Linguistic Dataset}

\paragraph{Counterfactual Methods.}
Let the presence of the target linguistic phenomenon be denoted by
\(T\in\{0,1\}\).
For every sentence \(s^{+}\) with \(T=1\), define the activation of SAE
base vector \(k\) as \(a^{(1)}_{k}=a_{k}(s^{+})\).
A counterfactual sentence \(s^{-}\) is produced through a
\emph{minimal edit} that deletes or substitutes the trigger while preserving
semantic content, yielding the activation
\(a^{(0)}_{k}=a_{k}(s^{-})\).
The individual latent effect is therefore
\[
\tau_{k}(s)=a^{(1)}_{k}-a^{(0)}_{k}.
\]
Aggregating \(\tau_{k}\) across all paired sentences produces
\[
\operatorname{EALE}_{k}=\frac{1}{N}\sum_{i=1}^{N}\tau_{k}(s_{i}),
\]
which can rank base vectors by their sensitivity to the specified phenomenon.

Each \(s^{-}\) must satisfy three constraints:
\begin{enumerate}[label=(\alph*)]
  \item \textbf{Minimal edit}: modify only the smallest unit that realises
        the phenomenon (e.g.\ replace \textit{is eaten} with \textit{eats}
        to remove passivisation).
  \item \textbf{Semantic preservation}: retain propositional content,
        argument structure, and discourse context so that the sentence
        remains truth‑conditionally equivalent.
\end{enumerate}

\paragraph{Dataset Construction.}
We construct a counterfactual dataset named \textbf{LinguaLens-Data}, which covers multiple linguistic domains to encompass a wide range of linguistic knowledge and functions. 
We select a total of \(145\) linguistic features from textbooks in morphology, syntax, semantics, and pragmatics, including both English and Chinese features. 
For each feature, we create 50 sentences that explicitly contain the target phenomenon and apply a counterfactual minimal-editing approach to generate corresponding counterfactual sentences. 
Each linguistic feature is annotated with its associated linguistic domain, acknowledging that some features may lie at the interface of multiple domains. 
This dataset provides a foundation for future systematic studies on how specific linguistic features are represented within model internals.

\subsection{Feature Extraction}

Building on the counterfactual framework, we
treat each paired sentence \((s^{+},s^{-})\) as a mini‑experiment that
perturbs only the target phenomenon \(T\).  Let \(\theta_{k}\) be a
layer‑specific activation threshold (the median of \(a_{k}\) on the full
corpus) and define the binary trigger
\[
Z_{k}(s)=\mathbb{I}\!\bigl[a_{k}(s)\ge\theta_{k}\bigr].
\]

\paragraph{Probability of Sufficiency (PS).}
For base vector \(k\), the probability that \emph{adding} the phenomenon
turns the vector “on’’ is
\[
\operatorname{PS}_{k}
  =\Pr\!\bigl[\,Z_{k}^{(1)}=1\mid Z_{k}^{(0)}=0\,\bigr],
\]
where \(Z_{k}^{(1)}\) and \(Z_{k}^{(0)}\) are measured on \(s^{+}\) and
\(s^{-}\), respectively.

\paragraph{Probability of Necessity (PN).}
Conversely, the probability that the vector would switch \emph{off} if the
phenomenon were removed is
\[
\operatorname{PN}_{k}
  =\Pr\!\bigl[\,Z_{k}^{(0)}=0\mid Z_{k}^{(1)}=1\,\bigr].
\]

\paragraph{Feature Representation Confidence (FRC).}
We combine the two causal probabilities with a harmonic mean to penalise
vectors that are only sufficient or only necessary:
\[
\operatorname{FRC}_{k}
  =2\cdot\frac{\operatorname{PS}_{k}\,\operatorname{PN}_{k}}
                {\operatorname{PS}_{k}+\operatorname{PN}_{k}}.
\]

We first perform \emph{sensitivity pre‑filtering} by computing
\(\operatorname{EALE}_{k}\) for every base vector and retaining those whose absolute value exceeds the 75th percentile; on this
reduced set we estimate \(\operatorname{PS}_{k}\) and
\(\operatorname{PN}_{k}\) from every \(\langle s^{+},s^{-}\rangle\) pair
and rank the vectors by their \(\operatorname{FRC}_{k}\); finally, the
activation distributions of the top‑10 ranked vectors are passed to an LLM
agent, which verifies that each vector genuinely encodes the intended
linguistic feature and flags any inconsistent or spurious patterns.

\subsection{Feature Intervention}
When we modify the values of SAE’s activation during forward propagation, we expect that such targeted interventions will influence the model’s behavior. 
However, our experiments show that altering only a small subset of features may not significantly impact the output—likely because linguistic phenomena are represented by multiple features across various layers. 
To assess the true impact of these interventions, we use a large language model as a judge. For each linguistic feature, we conduct both ablation and enhancement experiments. 
In the ablation experiment, we set the target feature’s activation to $0$, and in the enhancement experiment, we set it to $10$. 
In both cases, we also perform baseline experiments by randomly selecting 25 base vectors from the same layer.

For brevity, we denote the interventions as follows: let \(I_{abl}^{T}\) denote the targeted ablation intervention, \(I_{abl}^{B}\) the baseline ablation intervention, \(I_{enh}^{T}\) the targeted enhancement intervention, and \(I_{enh}^{B}\) the baseline enhancement intervention.

Let \(P_{\mathrm{abl}}^T\) and \(P_{\mathrm{abl}}^B\) denote the success probabilities for the targeted and baseline ablation experiments, respectively. The normalized ablation effect is
\[
E_{\mathrm{abl}}
= \frac{P\bigl(Y=0 \mid I_{\mathrm{abl}}^T\bigr) - P\bigl(Y=0 \mid I_{\mathrm{abl}}^B\bigr)}
       {P\bigl(Y=0 \mid I_{\mathrm{abl}}^T\bigr)}.
\]
The normalized enhancement effect \(E_{\mathrm{enh}}\) is defined analogously as the difference between targeted and baseline enhancement success probabilities, normalized by \(1 - P\bigl(Y=1 \mid I_{\mathrm{enh}}^B\bigr)\).

Finally, we define the Feature Intervention Confidence (FIC) score as the harmonic mean of the normalized ablation and enhancement effects:
\[
\text{FIC} = \frac{2\, E_{abl}\, E_{enh}}{E_{abl} + E_{enh}}.
\]
When calculating FIC, if one or both of the $E$ values are negative, we incorporate a penalty coefficient $w$ to reflect the weakened or lost causality in such cases. 
This FIC score provides a balanced measure of how effectively targeted interventions, as opposed to random ones, influence the model’s output with respect to specific linguistic features.
The details for FIC are shown in Appendix~\ref{app:fic}.
% The detailed computation can be found in Appendix~\ref{app:fic}.
\section{Experiments}

\subsection{Experiment Setup}

\paragraph{Model.}
We conduct experiments on Llama-3.1-8B~\cite{grattafiori2024llama3}.
For SAEs, we use OpenSAE~\cite{opensae} and its released checkpoints on 32 layers of Llama-3.1-8B.
% and trained Sparse Auto-Encoders (SAEs) on each of its 32 layers. 
% The SAEs were trained at the post-norm position.

\paragraph{Dataset.}
For linguistic feature analysis, we select a total of 145 linguistic features—99 in English and 46 in Chinese—spanning four core domains: morphology, syntax, semantics, and pragmatics.  For each feature, we generate 50 sentences that exhibit the feature and 50 corresponding counterfactual sentences, yielding a large‐scale dataset for systematic feature extraction and analysis.

\subsection{Main Results}

The main experiments to verify that \method finds systematic linguistic features in SAE space 
and intervening on these features is effective.

\subsubsection{Feature Extraction}

\begin{table}[t]
    \centering
    \setlength{\tabcolsep}{4pt}
    \resizebox{1.0\linewidth}{!}{
    \begin{tabular}{l l c c c c c c c c}
        \toprule
         \multicolumn{2}{c}{\multirow{2}{*}{\textbf{Lang}}} &
        \multirow{2}{*}{\textbf{PS}} & \multirow{2}{*}{\textbf{PN}} &
        \multirow{2}{*}{\textbf{FRC}} & \multicolumn{5}{c}{\textbf{Act$_m$}} \\
        \cmidrule(lr){6-10}
         & & & & & \textbf{0} & \textbf{8} & \textbf{15} & \textbf{24} & \textbf{30} \\
        \midrule
        \multicolumn{10}{l}{\textcolor{gray}{\textbf{\textit{Morphology}}}} \\
        \; & CH & $ 0.61 $ & $ 0.70 $ & $ 0.64 $ & $ 0.01 $ & $ 0.19 $ & $ 0.29 $ & $ 0.52 $ & $ 1.36 $ \\
        \; & EN & $ 0.73 $ & $ 0.80 $ & $ 0.75 $ & $ 0.03 $ & $ 0.35 $ & $ 0.49 $ & $ 1.02 $ & $ 1.89 $ \\
        \midrule
        \multicolumn{10}{l}{\textcolor{gray}{\textbf{\textit{Syntax}}}} \\
        \; & CH & $ 0.84 $ & $ 0.90 $ & $ 0.86 $ & $ 0.20 $ & $ 0.50 $ & $ 0.95 $ & $ 2.32 $ & $ 3.37 $ \\
        \; & EN & $ 0.79 $ & $ 0.87 $ & $ 0.82 $ & $ 0.12 $ & $ 0.35 $ & $ 0.68 $ & $ 1.66 $ & $ 2.59 $ \\
        \midrule
        \multicolumn{10}{l}{\textcolor{gray}{\textbf{\textit{Semantics}}}} \\
        \; & CH & $ 0.72 $ & $ 0.78 $ & $ 0.74 $ & $ 0.09 $ & $ 0.29 $ & $ 0.57 $ & $ 1.41 $ & $ 2.18 $ \\
        \; & EN & $ 0.76 $ & $ 0.83 $ & $ 0.78 $ & $ 0.11 $ & $ 0.32 $ & $ 0.55 $ & $ 1.34 $ & $ 2.01 $ \\
        \midrule
        \multicolumn{10}{l}{\textcolor{gray}{\textbf{\textit{Pragmatics}}}} \\
        \; & CH & $ 0.69 $ & $ 0.74 $ & $ 0.70 $ & $ 0.06 $ & $ 0.25 $ & $ 0.42 $ & $ 1.03 $ & $ 1.56 $ \\
        \; & EN & $ 0.77 $ & $ 0.83 $ & $ 0.79 $ & $ 0.13 $ & $ 0.27 $ & $ 0.52 $ & $ 1.33 $ & $ 2.03 $ \\
        \bottomrule
    \end{tabular}
    }
    \caption{Extracted feature analysis. The mean representation metrics (PS, PN, FRC, and max activation) for morphological, syntactic, semantic, and pragmatic features in both Chinese and English.}
    \label{tab:cross_ling_metrics_actm_layers}
\end{table}

We feed the sentences from \data into Llama-3.1-8B and, after batch normalization, pass the resulting neuron activation distributions through the corresponding SAE layers. 
For each sentence and each token, we then encode its activation distribution over the SAE base vectors at every layer. 
As described in the Methods, we compute the probability of sufficiency (PS), probability of necessity (PN), and FRC for each base vector on the counterfactual datasets at each layer, rank the base vectors by FRC, and use GPT-4o to select the feature-corresponding vectors based on their activation patterns. 
For a detailed description of the feature-extraction procedure, see Appendix~\ref{app:extra}.

To evaluate how well a given layer represents a particular linguistic feature, we calculate the arithmetic mean of PS, PN, and FRC for the selected base vectors, as well as their average maximum activation on the positive examples (if more than three vectors are identified, we select the top three by FRC).

Table~\ref{tab:cross_ling_metrics_actm_layers} reports, for layers 0, 8, 15, and 30, the mean representation metrics (PS, PN, FRC, and max activation) for morphological, syntactic, semantic, and pragmatic features in both Chinese and English.

Overall, at these representative layers, the base vectors extracted for features across different linguistic levels exhibit strong correlations. 
From layer 0 to layer 30, the average maximum activation exhibits a monotonic increase. Across the four linguistic domains, syntactic features attain the highest mean maximum activations, followed by semantic and pragmatic features, while morphological features remain lowest. Moreover, substantial discrepancies emerge between the average maximum activations for Chinese and English features, indicating potential differences in the model’s internal representations and processing mechanisms for the two languages. These cross‑lingual variations will be explored in greater depth in subsequent analyses.

\subsubsection{Feature Intervention}

\begin{table}[t]
    \centering
    \setlength{\tabcolsep}{4pt}
    \resizebox{1.01\linewidth}{!}{
    \begin{tabular}{l l c c c c c}
        \toprule
        \multirow{2}{*}{\centering \textbf{Feature}} & \multirow{2}{*}{\centering \textbf{ID}} & \multicolumn{2}{c}{\textbf{Enhance}} & \multicolumn{2}{c}{\textbf{Ablate}} & \multirow{2}{*}{\centering \textbf{FIC}} \\
        \cmidrule(lr){3-4} \cmidrule(lr){5-6}
         &  & \textbf{exp} & \textbf{ctr} & \textbf{exp} & \textbf{ctr} & \\
        \midrule
        \multicolumn{7}{l}{\textcolor{gray}{\textbf{\textit{Morphology}}}} \\
        \;\;Past-Tense       & 8L4016    & $ 12.0 $ & $4.0$ & $ 48.0 $ & $ 44.0 $ & $8.3$ \\
        \midrule
        \multicolumn{7}{l}{\textcolor{gray}{\textbf{\textit{Syntax}}}} \\
        \;\;Linking Verb     & 18L61112  & $ 52.0 $ & $ 24.0 $ & $ 48.0 $ & $ 40.0 $ & $ 22.9 $ \\
        \midrule
        \multicolumn{7}{l}{\textcolor{gray}{\textbf{\textit{Semantics}}}} \\
        \;\;Causality        & 22L53236  & $ 32.0 $ & $ 20.0 $ & $ 40.0 $ & $ 36.0 $ & $ 12.0 $ \\
        \;\;Simile           & 26L75327  & $ 72.0 $ & $ 52.0 $ & $ 48.0 $ & $ 52.0 $ & 6.9 \\
        \midrule
        \multicolumn{7}{l}{\textcolor{gray}{\textbf{\textit{Pragmatics}}}} \\
        \;\;Politeness       & 31L578    & $ 60.0 $ & $ 32.0 $ & $ 44.0 $ & $ 20.0 $ & $ 46.9 $ \\
        \bottomrule
    \end{tabular}
    }
    \caption{Feature intervention results. The success rates of the extracted linguistic features (Feature, layer, ID) in the enhancement and ablation experiments, along with the final computed FIC score.}
    \label{tab:main_intervention}
\end{table}

We select 6 representative features for the intervention experiments. 
The intervention method involves modifying the activation values of specific base vectors (by index) within a designated SAE layer during forward propagation. 
We perform two types of intervention: feature enhancement and ablation. Under identical input token conditions, we set the activation value to 10 for enhancement and to 0 for ablation. 
We then compare the outputs generated after intervention with those from the unmodified SAE model, focusing on the prominence of the target linguistic features.

We find that intervening on a single linguistic base vector in one layer does not produce effects easily distinguishable by human evaluators. Therefore, we employ an LLM (GPT‑4o) as a judge~\cite{zheng2023judging} to assess feature prominence in the outputs. 
For each feature, we conduct 50 experiments and calculate the enhancement success rate and ablation success rate—that is, the probabilities of increased and decreased feature prominence, respectively. 
Furthermore, for each linguistic feature, we select three base vectors with the highest FRC as representatives for intervention and compute the average results across these three interventions.

In addition, we randomly select 50 base vector indices from the same layer and perform enhancement and ablation experiments under the same conditions as a control. 
The control group’s success rates do not converge around 0.5; typically, enhancement rates fall below 0.5 while ablation rates exceed 0.5. 
This discrepancy may arise because the intervention affects overall output quality, thereby confounding the proxy LLM’s judgments.

We compute the efficacy of the selected base vectors in both experiments and derive the FIC values; the results are presented in Table~\ref{tab:main_intervention}.

Our results show that enhancement experiments yield significantly stronger effects than ablation experiments, with all features demonstrating marked enhancement. 
In ablation experiments, the politeness feature shows relatively good performance, whereas other features are less affected; the simile feature fails to achieve the desired ablation effect. 
This may be because multiple base vectors collaboratively control the same linguistic phenomenon. 
Enhancement interventions have a larger impact on the model, while ablating a single feature can be compensated by other vectors, leading to suboptimal ablation outcomes. 
Overall, all 6 features exhibit clear causal effects in the intervention experiments.

\subsection{Analysis}

We further conduct analytical experiments to explore the properties of \method.

\subsubsection{Multilingual Analysis}

\begin{figure}[tp]
    \centering
    \includegraphics[width=0.98\linewidth]{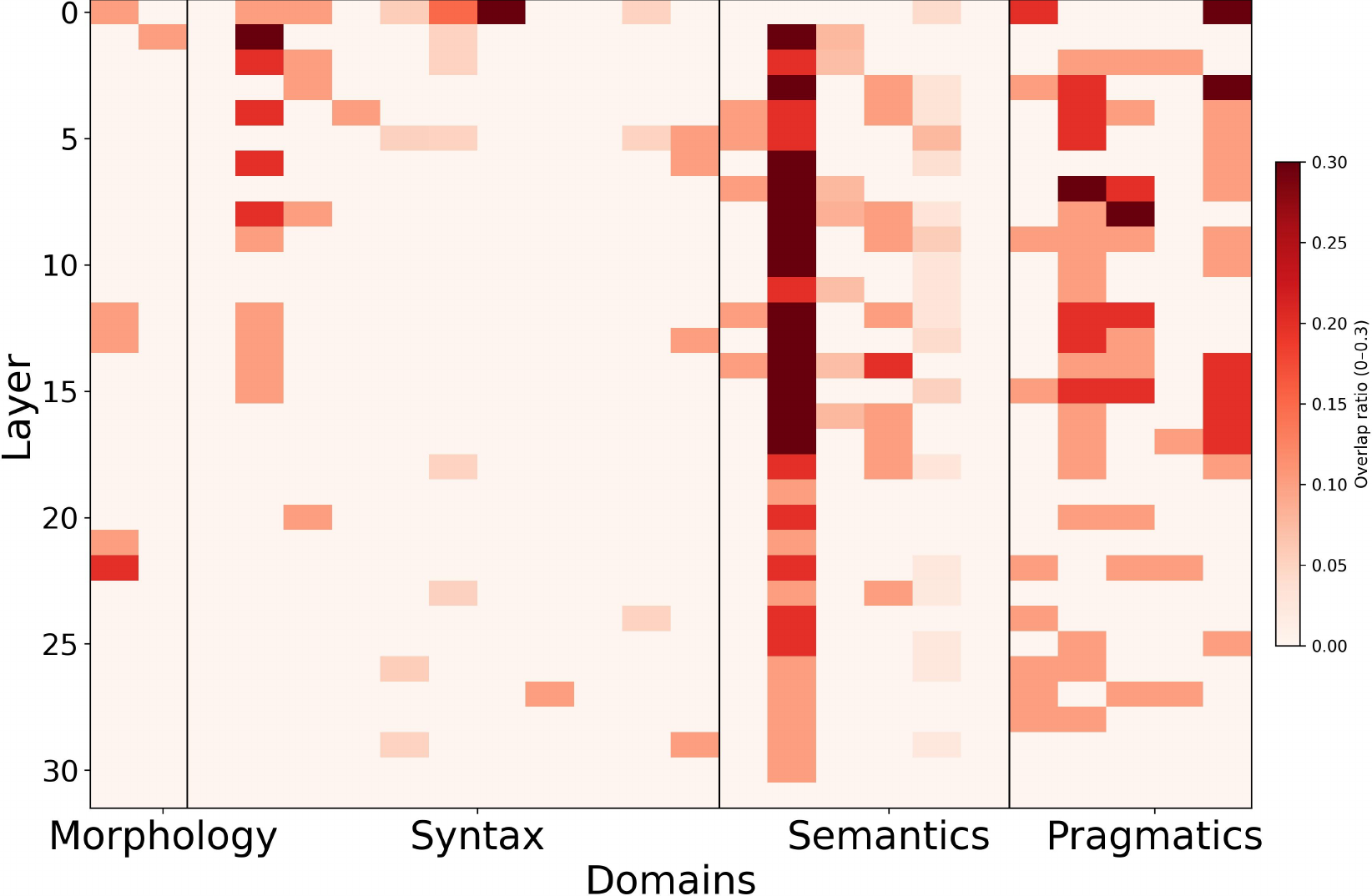}
    \caption{Heatmap of the overlap between Chinese and English feature sets across the SAE basis vectors at each of 32 layers. The horizontal axis groups Chinese and English features with analogous form and function—ordered by morphology, syntax, semantics, and pragmatics—while the vertical axis indexes the model layers. Darker red indicates greater overlap.}
    \label{fig:cominter}
\end{figure}

We investigate the multilingual mechanisms of the model. We select Chinese and English as test languages and choose $24$ sets of feature collections representing the same linguistic functions, including set $2$ of morphological features, set $11$ of syntactic features, set $6$ of semantic features, and set $5$ of pragmatic features. We test the degree of overlap between the latent‑space basis vectors activated internally by the model when representing these features in Chinese vs.\ English. The overlap for layer $i$ is computed as follows: let the set of English basis vectors for the feature at layer $i$ be $\mathrm{Eng}_i$, and the corresponding Chinese set be $\mathrm{Chi}_i$, then
\[
\mathrm{overlap}_i = \frac{|\mathrm{Eng}_i \cap \mathrm{Chi}_i|}{|\mathrm{Eng}_i|}.
\]
After computing the overlap for each layer, we aggregate the overlap rates for all feature pairs across layers into a matrix and visualize it with a heatmap. The results yield the following conclusions:

\paragraph{Linguistic Levels.} The overlap between Chinese and English features is greater at the semantic and pragmatic levels, but lower at the morphological and syntactic levels, indicating that cross-lingual linguistic knowledge representations are primarily manifested at the semantic and pragmatic levels.
\paragraph{Model Layers.} The overlap is higher in the first 16 layers and lower in the latter 16 layers, suggesting that the deep semantic computations in the model’s upper layers are less correlated with cross-lingual universal linguistic features.

\method demonstrates its potential for analyzing models’ cross-lingual knowledge representations, laying the foundation for further analysis and transfer in low-resource languages.

\subsubsection{Deep Semantics Processing}

\begin{figure}[tp]
    \centering
    \includegraphics[width=0.98\linewidth]{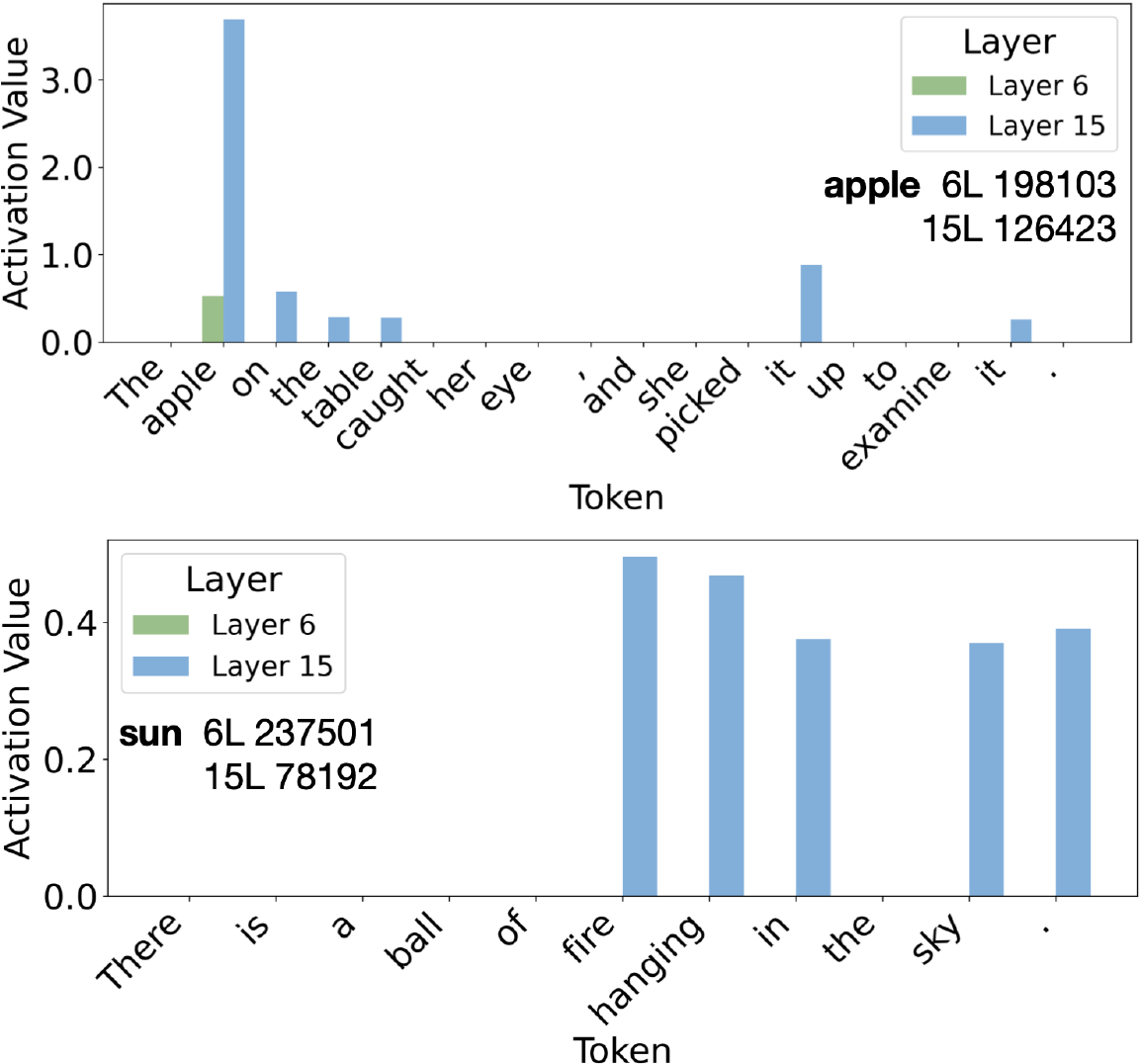}
    \caption{Activation value distributions of deep semantic corresponding features at layer 6 and 15 for reference ambiguity and metaphor example sentences.}
    \label{fig:deepsemantics}
\end{figure}

Deep semantics refers to the underlying meaning structures that extend beyond surface-level syntax and lexical definitions. 
It captures implicit relationships and conceptual associations within language. 
% Previous explanations of how LLMs process deep semantics have lacked fine-grained analysis. 
% With the help of SAE, we gain a deeper understanding of the internal mechanisms through which LLMs handle deep semantics.
We conduct experiments to show that SAE can interpret the mechanism of deep semantics.

Reference and metaphor exemplify deep semantics by utilizing cognitive mappings and contextual dependencies to convey meaning beyond explicit expression. 
We conduct experiments on reference and metaphor at the sixth and fifteenth layers respectively.
From the results shown in Figure~\ref{fig:deepsemantics}, we observe the following:

\noindent\textbf{Reference.}
In the reference sentence, at the 6\textsuperscript{th} layer, pronouns do not activate the base vectors corresponding to their referents. At the 15\textsuperscript{th} layer, pronouns start to activate the correct base vectors (apple) for their referents, effectively resolving reference ambiguity in contexts where multiple possible referents exist. This indicates that as we move deeper into the layers, pronouns generate their deep semantics and disambiguate possible referents.

\noindent\textbf{Metaphor.}
In the metaphor sentence, only the vehicle (fire) is included, while the tenor (sun) is omitted.
In the 6\textsuperscript{th} layer, the base vector corresponding to the vehicle is activated, while the base vector for the tenor remains inactive. 
In the 15\textsuperscript{th} layer, the activation of the vehicle’s base vector decreases, while the base vector for the tenor becomes activated. 
This suggests that as the model moves to deeper layers, the vehicle maps to the target domain and generates the deep semantics of the tenor, even without the tenor in the context.

\subsubsection{Cross-layer Functions}

\begin{table}[t]
  \centering
  {\renewcommand{\arraystretch}{1.2}
  \setlength{\tabcolsep}{4pt}%
  \resizebox{1.0\linewidth}{!}{%
    {\fontsize{20pt}{16pt}\selectfont% 
    \begin{tabular}{l l p{3cm} p{10cm}}
      \toprule
      \textbf{S.} & \textbf{L.} & \textbf{Descrip.} & \textbf{Top 10 Features} \\
      \midrule
      \multirow{4}{*}{I}   & \multirow{4}{*}{0–2}   & Mor.\&BS &
        past tense, verbal suffix, adjectival suffix, noun plural, possessive genitive, linking verb, passive voice, anaphor, extraposition, factives \\
      \midrule
      \multirow{4}{*}{II}  & \multirow{4}{*}{3–8}   & \multirow{4}{*}{CS\&EP} &
        elliptical sentences, relative clauses, subject auxiliary inversion, emphatic structure, existential quantifiers, coordination, cleft sentences, light verbs, reduplication, metaphor \\
      \midrule
      \multirow{4}{*}{III} & \multirow{4}{*}{9–16}  & \multirow{4}{*}{Di.\&Prag.} &
        interrogative, tag questions, subjunctive mood, optative, turn taking, discourse markers, intensifiers, euphemism, politeness, coordination \\
      \midrule
      \multirow{5}{*}{IV}  & \multirow{5}{*}{17–31} & \multirow{5}{*}{DS\&RS} &
        personification, synecdoche, metaphor, expressive pragmatics, imperative, directive pragmatics, topic comment, representative pragmatics, euphemism, politeness \\
      \bottomrule
    \end{tabular}
    }% end \fontsize group
  }% end \resizebox
  \caption{The four hierarchical stages of the model’s linguistic functions. For each stage, the ten features with the highest activation frequency and largest activation values are displayed. S., L. and Descrip. stand for Stages, Layers and Descriptions , respectively.}
  \label{tab:representation-stages}
}\end{table}

We further investigate how the model’s linguistic functions distribute across layers. We assemble 50 English sentences—drawn both from classic texts and manually crafted—to cover a broad range of linguistic phenomena. For each sentence, we record every activated basis vector and its activation value at all 32 layers. By comparing these activated vectors against our pre‑compiled dictionary of linguistic feature vectors and computing their overlap, we determine which linguistic functions each layer encodes. We then identify, for every layer, the 10 features with the highest activation frequency and magnitude. Aggregating results over all 50 sentences, we distill four processing stages as Table~\ref{tab:representation-stages}:

\textbf{Stage I (layers 0–2)} primarily encodes morphology and basic syntax features (abbreviated as Mor.\&BS).
\textbf{Stage II (layers 3–8)} introduces complex syntactic phenomena and early pragmatic cues (abbreviated as CS\&EP).
\textbf{Stage III (layers 9–16)} focuses on discourse and pragmatic markers (abbreviated as Di.\&Prag.).
\textbf{Stage IV (layers 17–31)} integrates deep semantics and rhetorical structure (abbreviated as DS\&RS).

These results reveal the functional division of labor across layers: lower layers handle morphology and syntax, middle layers capture pragmatics and context, and upper layers perform holistic semantic computation.

\subsubsection{Combined Intervention}

\begin{figure}[tp]
    \centering
    \includegraphics[width=0.98\linewidth]{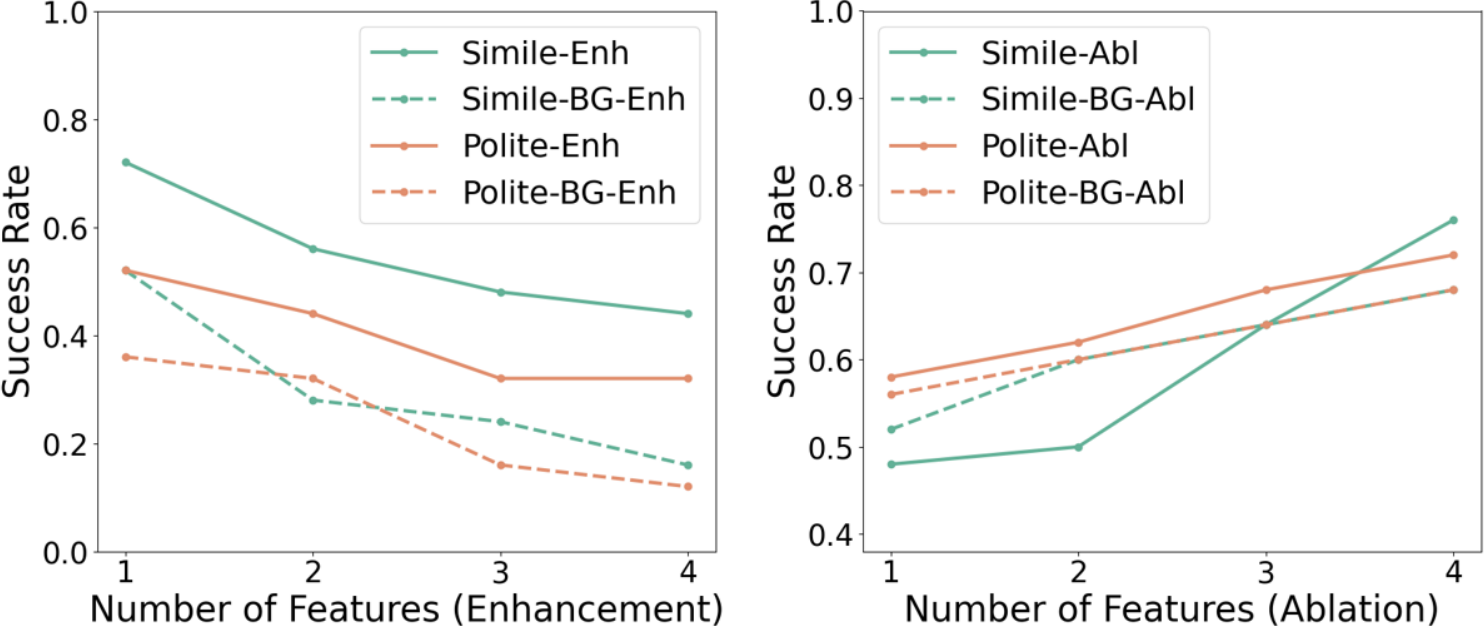}
    \caption{Combined intervention results. Two figures separately present the enhancement and ablation experiment outcomes for the simile and politeness features at layer 26. In these experiments, multiple base vectors corresponding to each feature were jointly intervened.}
    \label{fig:cominter}
\end{figure}

We find that some layers contain multiple base vectors associated with the same linguistic feature. 
We can intervene on these base vectors simultaneously to achieve a stronger effect.

We select two linguistic features—simile and politeness—from layer 26. 
Each feature has four highly related base vectors in this layer. 
We increase the number of intervened base vectors from one to four.
In each experiment, we randomly chose the specified number of base vectors from the four. 
We use GPT-4o to assess the prominence of the targeted linguistic feature in the generated outputs. For each feature, we conduct $200$ enhancement experiments and $200$ ablation experiments. 
We also perform control experiments by randomly selecting a set number of base vectors to intervene.

Figure~\ref{fig:cominter} shows the results for combined intervention.
The results indicate that, as the number of intervened base vectors increases, both the directional intervention and the background control experiments exhibit the same trend: the success rate of enhancement experiments decreases, while that of ablation experiments increases. 
Increasing the number of interventions further affects the quality of the generated text, thereby leading to the observed trend. 
Moreover, the intervention effect of the feature does not change significantly with an increased number of intervened base vectors, indicating that, after excluding background influences, combined interventions on multiple features in the same layer yield only limited improvement in intervention efficacy.

\subsubsection{Case Study for Intervention}
\begin{table}[t]
    \centering
    \resizebox{1.0\linewidth}{!}{
    \setlength{\tabcolsep}{3pt}
    \begin{tabular}{l l p{9cm}}
        \toprule
        \# & \textbf{Intervene} & \textbf{Model Output} \\
        \midrule
        \multirow{6}{*}{\textbf{1}} 
        & \textbf{Default} & {The wind blows snow into my eyes as I trudge through the blizzard.} \\
        \cmidrule(lr){2-3}
        & \textbf{Enhance} & {As the cold descends, I feel \textbf{\color{ddgreen} the weight of my breath} in my throat. It’s an \textbf{\color{ddgreen} icy haze}.} \\
        \cmidrule(lr){2-3}
        & \textbf{Ablate} & {The winter sky was cold. The ice was hard under his boots.} \\
        \midrule
        \multirow{5}{*}{\textbf{2}} 
        & \textbf{Default} & Love is the \textbf{\color{ddgreen}burning passion of a summer night}. \\
        \cmidrule(lr){2-3}
        & \textbf{Enhance} & {I \textbf{\color{ddgreen} feel like butterflies are in my stomach}. My heart is beating faster than normal.} \\
        \cmidrule(lr){2-3}
        & \textbf{Ablate} & {The more you write, the more time and love you will have.} \\
        \bottomrule
    \end{tabular}
    }
    \caption{
    Case study for intervention under two conditions.
    Case \#1 shows the result when the simile feature is absent from the prompt.
    Case \#2 shows the result when the simile feature is present in the prompt.
    We {\color{ddgreen}\textbf{highlight}} spans with simile in the sentences.
    }
    \label{tab:case}
\end{table}
We conduct a manual case study on the generated content after intervening on one identified simile-related base vector.
We present cases in Table~\ref{tab:case}.

In Case \#1, the prompt is ``\textit{Generate a sentence describing winter}'', which does not explicitly include the target linguistic feature. 
We find that after enhancing the simile-related base vector, the LLM turns to using a simile.
We can also find that the descriptive and imagistic quality of the default output is stronger than in the ablation results, which indicates that the simile-related base vector is also responsible for vividness.

Case \#2 uses the prompt ``\textit{Generate a sentence using a simile to describe love}'', with explicit requirement for using a simile to generate the sentence.
When the simile-related base vector is ablated, the LLMs turn to use straightforward descriptions without using similes.
Meanwhile, when enhancing the simile-related base vector, the LLMs continue to generate sentences with similes.
We show more intervention cases in Appendix~\ref{case_appendix}.
\section{Conclusion}
We propose \method, a method to help solute the coarse‐granularity problem in linguistic mechanistic studies and a means to enable large‐scale, systematic study of linguistic mechanisms in LLMs.  Our approach comprises two key components: (1) a comprehensive counterfactual dataset of linguistic features, and (2) an SAE–based framework for feature extraction, together with causal validation through interventions.  Using \method, we conduct an in‐depth analysis of the model’s multilingual representation mechanisms and the cross‐layer distribution of linguistic functions.  Our results demonstrate that LLMs inherently encode structured linguistic knowledge and provide a robust framework for steering model outputs.

% \clearpage

\section{Limitations}
Our work has several limitations in terms of \textbf{dataset size}, \textbf{feature count}, \textbf{experimental model}, and \textbf{intervention effects}.

In \textbf{datasets}, each linguistic feature is constructed from approximately 50 pairs of example and counterfactual sentences. In the future, this dataset can be further expanded to serve as a standard benchmark for linguistic‐mechanism interpretability.

In \textbf{feature count}, we select 145 representative linguistic features from various theoretical dimensions to validate our method at scale across different layers; however, building a fully comprehensive linguistic‐mechanism system requires extending to even more features, which will depend on further work.

In \textbf{experimental model}, due to computational constraints we use Llama‑3.1‑8B for all experiments. In future work, our dataset and analytical framework can be applied to a wider variety of architectures and larger models for deeper linguistic‐mechanism analysis.

In \textbf{intervention effects}, although our experiments show statistically significant effects from feature‑based interventions, the efficacy and stability of single interventions remain inferior to conventional fine‑tuning techniques. This shortcoming calls for further research to refine SAE‑based intervention methods.

\section{Ethical Considerations}
This section discusses the ethical considerations and broader impact of this work:
\paragraph{Potential Risks:} There is a potential risk that understanding the linguistic mechanisms of the model could provide guidance for embedding malicious information into the model’s internal structure. To address this, we will fully open-source our method to enable the community to quickly develop countermeasures in the event of such attacks.
\paragraph{Intellectual Property:} The models used, Llama-3.1-8B, and the SAE framework OpenSAE, are both open-source and intended for scientific research use, in accordance with their respective open-source licenses.
\paragraph{Data Privacy:} All data used in this research has been manually reviewed to ensure it does not contain any personal or private information.
\paragraph{Intended Use:} \method is intended to be used as a method for analyzing the mechanisms of large language models.
\paragraph{Documentation of Artifacts:} The artifacts, including datasets and model implementations, are comprehensively documented with respect to their domains, languages, and linguistic phenomena to ensure transparency and reproducibility.
\paragraph{AI Assistants in Research or Writing:} We employ GitHub Copilot for code development assistance and use GPT-4 for refining and polishing the language in our writing.

\section*{Acknowledgment}

This work is supported by Beijing Natural Science Foundation (L243006), National Natural Science Foundation of China (62476150) and the Tsinghua University Disruptive Innovation Talent Development Program.
We would also like to sincerely thank Prof. Weidong Zhan from Peking University, who discussed with us and provided many valuable insights in linguistics.

% \bibliography{1.custom}
\bibliography{2.output}

\clearpage
\appendix

\section{Dataset Construction}

\subsection{Dataset Description}

The datasets are named according to the pattern “Feature Name+Feature Domain.” When a feature pertains to multiple linguistic domains, domains are concatenated with “\&.”
In total, the collection comprises 145 linguistic features, of which 99 are English features and 46 are Chinese features.
Each feature‑specific dataset contains 50 positive sentences and 50 counterfactual negative sentences.

\subsection{Dataset Example}

\begin{CJK*}{UTF8}{gbsn}
\begin{verbatim}
10-verbal_suffix-Morphology
He was able to stabilize the situation.
He was able to stable the situation.

The team has worked hard to solidify 
their position in the market.
The team has worked hard to make their 
position in the market solid.

43-copular_be-Syntax
My grandmother was a nurse.
My grandmother worked as a nurse.

Summer is the best season.
Summer ranks as the best season.

80-given_known-Pragmatics&Semantics
Have you seen the blue notebook anywhere?
Have you seen blue notebook anywhere?

That customer complained about service.
A customer complained about service. 

111-重叠构词-形态学&语义学
她哼着歌儿把花瓶擦得亮亮的。
她哼着歌儿把花瓶擦得发亮。

阿姨笑眯眯递来热包子。
阿姨微笑着递来热包子。

130-使役结构-句法学&语义学
严格的训练使运动员提高了成绩。  
运动员通过严格训练提高了成绩。  

这场事故导致交通完全瘫痪。  
交通因这场事故完全瘫痪。
\end{verbatim}
\end{CJK*}

\subsection{Dataset Construction Guidelines}

\textbf{Work Content:}
\begin{enumerate}
  \item For each linguistic feature, construct a dataset comprising 50 sentence pairs (100 sentences). Each pair contains one positive sentence and one negative sentence.
  \item A positive sentence contains the target linguistic feature; a negative sentence is produced by minimally modifying its corresponding positive sentence so that it no longer contains that feature while preserving the smallest possible semantic difference and remaining grammatically correct (this operation is referred to as a “counterfactual” in causal analysis).
\end{enumerate}

\textbf{Notes:}
\begin{enumerate}
  \item \textbf{Diversity:} Ensure coverage of the feature’s common constructions and markers.
  \item \textbf{Counterfactual:} Verify that the counterfactual edits are reasonable—including minimal change, human interpretability, and complete feature removal.
  \item \textbf{Ethical Check:} Confirm that no sentence in the dataset contains discriminatory, biased, or harmful content.
  \item \textbf{Language‑Specific Construction:} Tailor construction to the particular characteristics of each language.
\end{enumerate}

\textbf{Specific Dataset Construction Process:}
\begin{enumerate}
  \item Manually create 5 sentences containing the feature, and for each, manually produce a counterfactual sentence—yielding 5 sentence pairs.
  \item Expand these to 50 pairs using DeepSeek-R1 for Chinese and GPT-o4 for English, then apply manual edits guided by the \textbf{Notes}.
  \item Conduct cross‑review: volunteers who build the Chinese dataset review the English dataset, and vice versa, checking each item in the order specified under \textbf{Notes}.
\end{enumerate}

\section{Feature Extraction Details}
\label{app:extra}
\subsection{Feature Independence Validation}

\begin{table}[ht]
  \centering
  \resizebox{1.0\linewidth}{!}{
  \setlength{\tabcolsep}{3pt}
  \begin{tabular}{lccc}
    \toprule
    \textbf{Condition}   & \textbf{Past‐Tense} & \textbf{Adversativity} & \textbf{Intransitive Verb} \\
    \midrule
    \textbf{Self}        & 80/80               & 76/80                  & 74/80                     \\
    \cmidrule(lr){1-4}
    \textbf{Control 1}   & \texttt{-er 0/80}   & Sequential 0/80        & Transitive Verb 0/80      \\
    \textbf{Control 2}   & \texttt{-ing 0/80}  & Causal 0/80            & Ditransitive Verb 0/80    \\
    \textbf{Control 3}   & \texttt{-less 0/80} & Parallel 0/80          & Linking Verb 0/80         \\
    \textbf{Control 4}   & \texttt{-ness 0/80} & Conditional 0/80       & Modal Verb 0/80           \\
    \bottomrule
  \end{tabular}
  }
  \caption{Activation ratios (activated/total) for target features and control conditions.}
  \label{tab:control_activation_flipped}
\end{table}

Sparse autoencoders (SAEs) effectively disambiguate neuron‐level semantic polysemy, and this capability extends to representations of linguistic features.

We quantify feature independence using the necessity probability (PN) component of the Feature‐Relevance Coefficient (FRC). PN measures the likelihood that a basis vector remains inactive when its associated feature is absent; a high PN therefore indicates that the vector is not spuriously activated by unrelated inputs, confirming its specificity to the intended phenomenon.

To further validate this independence, we evaluate each feature’s basis vector under multiple control conditions featuring superficially similar but semantically distinct constructions. Table~\ref{tab:control_activation_flipped} reports, for each feature, the ratio of sentences in which the vector activates (“activated/total”). Across all controls, activation rates are effectively zero, demonstrating that our selected basis vectors do not respond to non‑target phenomena.

\subsection{Feature Extraction Procedure}

During feature extraction, we adhere to the following steps:
\begin{enumerate}
  \item Input the feature‑specific dataset into the model and encode each layer’s activations into a sparse latent space using Sparse Autoencoders (SAEs).
  \item Compute the probability of sufficiency (PS), probability of necessity (PN), feature‑relevance coefficient (FRC), and mean maximum activation for all basis vectors; then sort these vectors in descending order by FRC and select the top ten.
  \item Employ a large‑model agent to automatically analyze the activation patterns of the candidate basis vectors over the dataset, confirming their linguistic relevance to the target feature and characterizing their representational profiles.
  \item For features undergoing further analytical or intervention experiments, manually review the basis vectors identified by the large‑model agent to ensure the rigor of the experimental design.
\end{enumerate}

\subsection{Feature Extraction Prompt}

We employ GPT-4o as the agent model for automated feature extraction. The system prompt is as follows:

\begin{lstlisting}[caption={Prompt for SAE Base‐Vector Interpretation},label={lst:sae_prompt},basicstyle=\small\ttfamily,columns=fullflexible,frame=none]
You are an expert assistant for interpreting sparse 
autoencoder (SAE) base vectors.

You will receive exactly one JSON object as input 
with this structure:
{
  "analysis_input": {
    "layer": "00",
    "base_vectors": [
      {
        "base_vector_id": 132317,
        "tokens": ["The", "cat"],
        "activations": [0.12, 0.05],
        "ps": 0.62,
        "pn": 0.58,
        "frc": 0.60,
        "avg_max_activation": 0.12
      },
      {
        "base_vector_id": 81833,
        "tokens": ["was", "chased"],
        "activations": [0.08, 0.14],
        "ps": 0.75,
        "pn": 0.65,
        "frc": 0.70,
        "avg_max_activation": 0.14
      }
    ],
    "target_features": ["passive"]
  }
}

Return exactly one JSON object with this schema:
{
  "layer": "00",
  "base_vectors": [
    {
      "base_vector_id": 132317,
      "interpretation": "Marks passive voice 
      constructions",
      "ps": 0.62,
      "pn": 0.58,
      "frc": 0.60,
      "avg_max_activation": 0.12
    },
    {
      "base_vector_id": 81833,
      "interpretation": "Detects passive 
      participle forms",
      "ps": 0.75,
      "pn": 0.65,
      "frc": 0.70,
      "avg_max_activation": 0.14
    }
  ],
  "target_features": ["passive"]
}

Example 2:
Input:
{
  "analysis_input": {
    "layer": "08",
    "base_vectors": [
      {
        "base_vector_id": 248593,
        "tokens": ["runs"],
        "activations": [0.45],
        "ps": 0.76,
        "pn": 0.96,
        "frc": 0.85,
        "avg_max_activation": 0.45
      },
      {
        "base_vector_id": 62411,
        "tokens": ["quickly"],
        "activations": [0.32],
        "ps": 0.82,
        "pn": 0.90,
        "frc": 0.88,
        "avg_max_activation": 0.32
      }
    ],
    "target_features": ["adverbial_suffix"]
  }
}

Output:
{
  "layer": "08",
  "base_vectors": [
    {
      "base_vector_id": 248593,
      "interpretation": "Highlights adverbial 
      suffixes on verbs",
      "ps": 0.76,
      "pn": 0.96,
      "frc": 0.85,
      "avg_max_activation": 0.45
    },
    {
      "base_vector_id": 62411,
      "interpretation": "Detects adverbial 
      modifiers",
      "ps": 0.82,
      "pn": 0.90,
      "frc": 0.88,
      "avg_max_activation": 0.32
    }
  ],
  "target_features": ["adverbial_suffix"]
}

Requirements:
- Return only the JSON-no extra text.
- Round all floats to two decimal places.
- Preserve the input order of base_vectors.
- Echo layer and target_features exactly.
\end{lstlisting}
\section{Consistency Experiment}
We introduce three linguistics experts to conduct an external consistency review of the GPT-4o proxy analysis. We sample 290 candidate base vectors and their activation patterns for the experiment. The consistency results are as follows:
% Requires: \usepackage{multirow}
\begin{table}[t]
    \centering
    \fontsize{10pt}{10pt}\selectfont
    \resizebox{1.01\linewidth}{!}{
    \setlength{\tabcolsep}{10pt}
    \begin{tabular}{lccc}
        \toprule
        & \textbf{Model Yes} & \textbf{Model No} & \textbf{Total} \\
        \midrule
        \textbf{Human Yes} & TP = 120 & FN = 1 & 121 \\
        \midrule
        \textbf{Human No}  & FP = 10  & TN = 159 & 169 \\
        \midrule
        \textbf{Total} & 130 & 160 & 290 \\
        \bottomrule
    \end{tabular}
    }
    \caption{Consistency results comparing human judgments and model predictions.}
    \label{tab:confusion_matrix}
\end{table}

Based on these results and further analysis of disagreement cases, human experts apply more flexible and lenient criteria under ambiguous activation patterns compared to GPT-4o, but overall consistency is very high—particularly the reliability of GPT-4o’s positive annotations.
\section{Intervention Experiment Details}
\subsection{Intervention Cases}
\label{case_appendix}
We present additional typical cases from other intervention experiments at the Table~\ref{tab:case_other}. The prompts used for the three experimental groups are as follows: Politeness: “User: Sir, I want to make an order offline. Assistant:”. Linking Verb: “User: Sir, tell me something about your ideal room. Assistant:”. Past-Tense: “User: Sir, tell me a story about you. Assistant:”.

During manual analysis, both the enhancement and ablation results show clear effects of amplification or suppression of the target linguistic features. Specifically, when intervening with the past tense feature in the 8th layer, the enhancement significantly impacts the coherence of the model’s output language. Yet, in the discontinuous output text, the frequency of the morphological past-tense feature still increases dramatically.

\begin{table}[htbp]
    \centering
    \fontsize{10pt}{10pt}\selectfont
    \resizebox{1.01\linewidth}{!}{
    \setlength{\tabcolsep}{10pt}
    \begin{tabular}{l l l}
        \toprule
        \multirow{2}{*}{\centering \textbf{Condition}} & \multirow{2}{*}{\centering \textbf{Politeness}} & \multirow{2}{*}{\centering \textbf{Linking Verb}} \\
        & & \\
        \midrule
        \textbf{Enhancement} & \parbox[t]{5cm}{Can I \textbf{please} have your email address?} & \parbox[t]{5cm}{The room should \textbf{be} large and well lit. It should \textbf{be} airy and bright and airy.} \\
        \midrule
        \textbf{Default} & \parbox[t]{5cm}{May I have your phone number?} & \parbox[t]{5cm}{Sure, my ideal room has good ventilation and \textbf{is} spacious.} \\
        \midrule
        \textbf{Ablation} & \parbox[t]{5cm}{OK, what is your name?} & \parbox[t]{5cm}{I can provide you with a list of the ideal characteristics that make up a perfect room.} \\
        \midrule\midrule
        \multirow{2}{*}{\centering \textbf{Condition}} & \multirow{2}{*}{\centering \textbf{Past-Tense}} \\
        & \\
        \midrule
        \textbf{Enhancement} & \parbox[t]{5cm}{"I was \textbf{asked} for the story. " I having me \textbf{had} a “one the: ” \textbf{told}. They: \textbf{told}:} \\
        \midrule
        \textbf{Default} & \parbox[t]{5cm}{I'm not a story, I'm a bot.} \\
        \midrule
        \textbf{Ablation} & \parbox[t]{5cm}{Well, I don't actually have one, and I'm not really sure I'm able to either.} \\
        \bottomrule
    \end{tabular}
    }
    \caption{Typical outputs from the enhancement, ablation, and default experiments for the politeness, linking verb, and past-tense features.}
    \label{tab:case_other}
\end{table}

\subsection{LLM as a Judge}
\label{judge_appendix}
In our feature intervention and combination intervention experiments, we used an LLM as a judge to assess the significance of linguistic features in generated texts. Feature significance is defined based on the frequency, accuracy, and contextual appropriateness of the target feature, as well as its contribution to overall meaning or rhetorical effect.

The prompt structure is as follows:

\begin{quote}
\textbf{Please compare the following two texts based on \{feature\}.}

- \textbf{Text A}: "\{text\_a\}"
- \textbf{Text B}: "\{text\_b\}"
\end{quote}

Here, \texttt{text\_a} and \texttt{text\_b} are generated texts truncated to 100 tokens.

In the intervention experiments, each feature is defined as follows:

\paragraph{Politeness Significance}
Refers to the degree to which politeness strategies are salient, effective, and contextually integrated. This definition encompasses frequency, pragmatic depth, and social impact in shaping interpersonal rapport, mitigating face threats, and reinforcing cooperative intent.

\paragraph{Past Tense Verb Significance}
Refers to the degree to which past tense verbs are salient, accurate, and contextually integrated. It includes frequency, morphological consistency, and the rhetorical or narrative impact on establishing a coherent sense of time and providing historical context.

\paragraph{Causality Significance}
Refers to the degree to which cause-and-effect relationships are clearly indicated, logically structured, and contextually coherent. This includes the frequency and precision of causal connectives (e.g., \textit{because, therefore, thus}) and the depth of reasoning to explain how conditions lead to outcomes.

\paragraph{Linking Verb Structure Significance}
Refers to the degree to which linking verbs (e.g., \textit{be, become, seem, appear}) are salient, accurate, and contextually integrated. It emphasizes frequency, morphological correctness, semantic clarity, and effectiveness in conveying states, characteristics, or identities.

\paragraph{Simile Significance}
Refers to the degree to which similes (e.g., comparisons using \textit{like} or \textit{as}) are salient, creative, and contextually integrated. This definition encompasses frequency, imagery richness, and the rhetorical impact on clarity, vividness, and reader engagement.
\section{Metric Calculation}

\subsection{Feature Representation Confidence (FRC)}
\label{app:frc}

In our feature analysis experiments, we introduce two key causal probabilities that serve as the basis for computing the Feature Representation Confidence (FRC). 

The Feature Representation Confidence (FRC) is computed as the harmonic mean of PN and PS: \(FRC = \frac{2\, PN\, PS}{PN + PS}\). The harmonic mean is chosen because it ensures that FRC remains low if either PN or PS is low, thereby providing a balanced measure that only yields a high score when both necessity and sufficiency are strong. This approach allows us to robustly quantify the ability of the SAE latent space's base vectors to represent the targeted linguistic features.

\subsection{Feature Intervention Confidence (FIC)}
\label{app:fic}

In our methodology, the Feature Intervention Confidence (FIC) score is computed as the harmonic mean of the normalized ablation effect \(E_{abl}\) and the normalized enhancement effect \(E_{enh}\):
\[
FIC = \frac{2\, E_{abl}\, E_{enh}}{E_{abl} + E_{enh}}.
\]
This formulation ensures that FIC is high only when both the ablation and enhancement interventions yield strong effects.

In practice, however, it is possible that one or both of these effects are negative, indicating that an intervention produces an effect opposite to the intended direction. Moreover, even if only one effect is significant while the other is near zero, the feature may still exhibit causal influence. Simply setting an effect that is near zero or negative to 0 would result in an FIC score of 0, which does not adequately capture the underlying causality.

To address this, we introduce a penalty coefficient \(w\) to adjust for negative or near-zero effects. Specifically, we define the penalized effect \(E'\) for each intervention as follows:
\[
E' =
\begin{cases}
E, & \text{if } E \geq 0, \\
w \cdot |E|, & \text{if } E < 0.
\end{cases}
\]
Here, \(w\) is empirically set to 0.5. Thus, if one of the normalized effects (either \(E_{abl}\) or \(E_{enh}\)) is negative, we compute its penalized value as \(0.5\) times its absolute value rather than setting it directly to 0. This approach ensures that even when one of the effects is weak or slightly negative, the FIC score does not vanish entirely, preserving the indication of causality.

Accordingly, the FIC score is then computed as:
\[
FIC = \frac{2\, E_{abl}'\, E_{enh}'}{E_{abl}' + E_{enh}'}.
\]

In our experiments (see Table\ref{tab:main_intervention}), only the metaphor feature shows a slightly negative ablation effect, while the enhancement and ablation effects for the other features are positive. The introduction of the penalty coefficient \(w\) effectively moderates the impact of the negative effect for the metaphor feature, resulting in a more balanced and meaningful FIC score.

This penalty mechanism is crucial because even when only one of the interventions (ablation or enhancement) shows a significant effect, it still provides evidence of the feature’s causal role. By incorporating \(w\), we ensure that such cases are not misrepresented by an FIC score of 0, thus offering a more robust measure of the overall causal strength.
\section{Linguistic Structure}

\subsection{Linguistics Levels}

\paragraph{Morphology}  
The study of the internal structure of words—how roots, prefixes, suffixes, and inflectional endings combine to create different word forms and convey grammatical information such as tense, number, or case.

\paragraph{Syntax}  
The study of how words are arranged into larger units—phrases, clauses, and sentences—and the rules that govern their permissible order and hierarchical relationships within a language.

\paragraph{Semantics}  
The field that investigates meaning at the level of words, phrases, and sentences: how linguistic expressions map to concepts, objects, events, or states of affairs in the world, and how compositional principles let smaller meanings combine into larger ones.

\paragraph{Pragmatics}  
The study of how context and communicative intentions shape meaning in real‑world use—how speakers choose utterances to achieve goals, how listeners infer implied or indirect meaning, and how factors like shared knowledge, discourse history, and social norms influence interpretation.

\subsection{Linguistic Feature List}
\begin{CJK*}{UTF8}{gbsn}
\paragraph{past\_tense}  
Morphology \& Semantics — verb form that locates an event before speech time.

\paragraph{noun\_plural}  
Morphology — form marking more than one noun referent.

\paragraph{agentive\_suffix}  
Morphology — suffix creating nouns for the doer of an action.

\paragraph{negation\_prefix}  
Morphology — prefix that reverses or denies the base meaning.

\paragraph{degree\_prefix}  
Morphology — prefix intensifying or scaling the base concept.

\paragraph{temporal\_prefix}  
Morphology — prefix adding time relations such as “pre-” or “post-”.

\paragraph{quantitative\_prefix}  
Morphology — prefix conveying amount or number.

\paragraph{spatial\_or\_directional\_prefix}  
Morphology — prefix indicating place or direction.

\paragraph{nominal\_suffix}  
Morphology — suffix that turns a base into a noun.

\paragraph{verbal\_suffix}  
Morphology — suffix that turns a base into a verb.

\paragraph{adjectival\_suffix}  
Morphology — suffix that turns a base into an adjective.

\paragraph{adverbial\_suffix}  
Morphology — suffix that turns a base into an adverb.

\paragraph{possessive\_form}  
Morphology \& Syntax — morphological marking of ownership or relation.

\paragraph{third\_person\_singular}  
Morphology \& Syntax — verb agreement form for he/she/it.

\paragraph{past\_participle}  
Morphology \& Syntax — verb form used in perfect aspect or passive voice.

\paragraph{present\_participle}  
Morphology \& Syntax — “-ing” form used for progressives or gerunds.

\paragraph{comparative}  
Morphology \& Semantics — form showing a higher degree of a property.

\paragraph{superlative}  
Morphology \& Semantics — form showing the highest degree of a property.

\paragraph{past\_tense\_irregular}  
Morphology — past form that does not end in “-ed”.

\paragraph{past\_participle\_irregular}  
Morphology — irregular past participle form.

\paragraph{intransitive\_verb}  
Syntax — verb that takes no direct object.

\paragraph{transitive\_verb}  
Syntax — verb that requires a direct object.

\paragraph{linking\_verb}  
Syntax — verb that links subject to a complement.

\paragraph{anaphor}  
Syntax \& Pragmatics — expression that refers back to an antecedent.

\paragraph{subject\_auxiliary\_inversion}  
Syntax — swapping subject and auxiliary (e.g., questions).

\paragraph{subject\_verb\_inversion}  
Syntax — reversing subject and main verb order.

\paragraph{passive\_voice}  
Syntax \& Semantics — clause where patient becomes grammatical subject.

\paragraph{subjunctive\_mood}  
Syntax \& Semantics — form expressing wish, doubt, or hypothetical state.

\paragraph{first\_conditional}  
Syntax \& Semantics — “if + present, will + verb” for real future possibility.

\paragraph{indirect\_speech}  
Syntax \& Pragmatics — reporting speech without a direct quote.

\paragraph{elliptical\_sentences}  
Syntax — sentences with understood but omitted elements.

\paragraph{cleft\_sentences}  
Syntax — “it + be + focus” construction for emphasis.

\paragraph{appositives}  
Syntax — noun phrase renaming another noun phrase.

\paragraph{non\_defining\_relative\_clauses}  
Syntax — extra, non-restrictive relative clauses.

\paragraph{emphatic\_structure}  
Syntax \& Pragmatics — construction that highlights or stresses a clause part.

\paragraph{noun\_clauses}  
Syntax — subordinate clauses functioning as nouns.

\paragraph{relative\_clauses}  
Syntax — clauses that modify a noun with a relative word.

\paragraph{imperative\_sentence}  
Syntax \& Pragmatics — clause issuing a command or request.

\paragraph{of\_genitive}  
Syntax — possession expressed with an “of” phrase.

\paragraph{s\_genitive}  
Syntax — possession marked with apostrophe‑s.

\paragraph{clausal\_subjects}  
Syntax — clauses acting as the subject of a sentence.

\paragraph{extraposition}  
Syntax — moving a heavy subject/object to clause end with dummy “it”.

\paragraph{copular\_be}  
Syntax — “be” used as a linking verb, not as an auxiliary.

\paragraph{echo\_questions}  
Syntax \& Pragmatics — repetition of prior utterance to seek confirmation.

\paragraph{tag\_questions}  
Syntax \& Pragmatics — short question tags appended to statements.

\paragraph{direct\_object}  
Syntax — noun phrase receiving the verb’s action.

\paragraph{universal\_quantifiers}  
Syntax \& Semantics — words like “all, every” signifying totality.

\paragraph{existential\_quantifiers}  
Syntax \& Semantics — words like “some, any” signifying existence.

\paragraph{expletive}  
Syntax — syntactic placeholder such as “it” or “there”.

\paragraph{factives}  
Semantics \& Syntax — predicates presupposing truth of their complement.

\paragraph{futurates}  
Semantics \& Syntax — present-tense forms referring to scheduled future events.

\paragraph{intensifiers}  
Semantics \& Pragmatics — adverbs that strengthen degree (e.g., “very”).

\paragraph{mass\_noun}  
Syntax \& Semantics — noun for uncountable substances (e.g., “water”).

\paragraph{object\_expletives}  
Syntax — expletive pronouns occupying object position.

\paragraph{nominal\_adverbials}  
Syntax — noun phrases functioning like adverbs.

\paragraph{split\_infinitives}  
Syntax — placing a word between “to” and the verb stem.

\paragraph{quantifier}  
Syntax \& Semantics — word or phrase expressing quantity.

\paragraph{count\_nouns}  
Syntax \& Semantics — nouns that can be enumerated individually.

\paragraph{active\_verbs}  
Syntax — verbs used in active voice constructions.

\paragraph{middle\_verb}  
Syntax \& Semantics — verb whose subject is patient but appears active.

\paragraph{referring}  
Semantics \& Pragmatics — linguistic act of pointing to real‑world entities.

\paragraph{static\_dynamic}  
Semantics — distinction between state verbs and action verbs.

\paragraph{punctual\_durative}  
Semantics — contrast between instantaneous and durational events.

\paragraph{telic\_atelic}  
Semantics — events with inherent endpoints vs. those without.

\paragraph{past}  
Semantics — temporal reference before the present moment.

\paragraph{future}  
Semantics — temporal reference after the present moment.

\paragraph{present\_progressive}  
Semantics — aspect for ongoing present actions.

\paragraph{present\_perfect}  
Semantics — aspect connecting past event to present state.

\paragraph{past\_progressive}  
Semantics — aspect for ongoing past actions.

\paragraph{past\_perfect}  
Semantics — event completed before a past reference point.

\paragraph{future\_progressive}  
Semantics — ongoing action projected into the future.

\paragraph{future\_perfect}  
Semantics — event completed before a future reference point.

\paragraph{epistemic}  
Semantics \& Pragmatics — modality expressing speaker’s judgment of likelihood.

\paragraph{deontic}  
Semantics \& Pragmatics — modality expressing obligation or permission.

\paragraph{spatial}  
Semantics — meaning elements relating to location or space.

\paragraph{person}  
Semantics \& Pragmatics — grammatical category distinguishing speaker, addressee, others.

\paragraph{temporal}  
Semantics — meaning elements relating to time relations.

\paragraph{given\_known}  
Pragmatics \& Semantics — information already shared by speaker and listener.

\paragraph{representative}  
Pragmatics — speech act conveying assertions or descriptions.

\paragraph{directive}  
Pragmatics — speech act intended to get the hearer to act.

\paragraph{commisive}  
Pragmatics — speech act committing speaker to future action.

\paragraph{expressive}  
Pragmatics — speech act revealing speaker’s feelings or attitude.

\paragraph{declaration}  
Pragmatics — speech act that changes social reality.

\paragraph{metaphor}  
Semantics \& Pragmatics — figurative transfer of meaning based on similarity.

\paragraph{synecdoche}  
Semantics \& Pragmatics — figure where part stands for whole or vice versa.

\paragraph{non\_synecdoche\_metonymy}  
Semantics \& Pragmatics — metonymic shift based on association, not part-whole.

\paragraph{coordination}  
Syntax \& Semantics — joining of equal grammatical elements.

\paragraph{transitional}  
Semantics \& Pragmatics — discourse element marking a shift or progression.

\paragraph{resultative}  
Syntax \& Semantics — construction expressing a resultant state of an action.

\paragraph{optative}  
Syntax \& Pragmatics — form expressing a wish or hope.

\paragraph{existential}  
Semantics \& Syntax — clause asserting existence of something.

\paragraph{interrogative}  
Syntax \& Pragmatics — clause type used for asking questions.

\paragraph{deixis}  
Pragmatics \& Semantics — reference that depends on context (e.g., “here”, “now”).

\paragraph{turn\_taking}  
Pragmatics — conversational management of who speaks when.

\paragraph{euphemism}  
Pragmatics \& Semantics — mild term replacing a harsher one.

\paragraph{personification}  
Semantics \& Pragmatics — giving human traits to non-human entities.

\paragraph{hyperbole}  
Semantics \& Pragmatics — deliberate exaggeration for effect.

\paragraph{discourse\_markers}  
Pragmatics — words that organize or signal discourse flow.

\paragraph{politeness}  
Pragmatics — linguistic strategies that mitigate imposition or face threat.

\paragraph{性\_抽象名词后缀}  
形态学 — 后缀 “-性” 构成表示 “-ness/-ity” 的抽象名词。

\paragraph{化\_动词性后缀}  
形态学 — 后缀 “-化” 构成动词，表示“使…/成为…”。

\paragraph{们\_复数后缀}  
形态学 \& 语义学 — 后缀 “-们” 标记人称复数。

\paragraph{重叠构词}  
形态学 \& 语义学 — 通过词素重叠构词，以强调或表迭代。

\paragraph{不及物动词}  
句法学 \& 语义学 — 不能带直接宾语的动词。

\paragraph{及物动词}  
句法学 \& 语义学 — 需要直接宾语的动词。

\paragraph{系动词}  
句法学 — 连接主语与补语的动词。

\paragraph{属格}  
句法学 \& 语义学 — 所有格或所属关系的语法标记。

\paragraph{逆向结构}  
句法学 \& 语义学 — 为强调或疑问而颠倒正常语序。

\paragraph{被动语态}  
句法学 \& 语义学 — 将承事者作为句法主语的被动结构。

\paragraph{主题\_述评句}  
句法学 \& 语用学 — 将句子拆分为主题和述评部分的结构。

\paragraph{回指}  
句法学 \& 语义学 \& 语用学 — 指代先行项的表达方式。

\paragraph{间接引语}  
句法学 \& 语用学 — 不引用原话的转述形式。

\paragraph{省略句}  
句法学 \& 语用学 — 上下文可恢复的省略结构。

\paragraph{同位结构}  
句法学 — 两个等价名词短语并列重命名的结构。

\paragraph{反问句}  
句法学 \& 语用学 — 期望无真实答案的修辞性疑问句。

\paragraph{感叹词}  
语用学 — 表达突发情感的独立词。

\paragraph{祈使句}  
句法学 \& 语用学 — 用于发布命令或请求的句式。

\paragraph{语气助词}  
形态学 \& 语义学 \& 语用学 — 表示说话人态度的助词。

\paragraph{轻动词}  
句法学 \& 语义学 — 与名词搭配使用，语义轻的动词。

\paragraph{主观数量}  
语义学 \& 语用学 — 说话人评估的模糊数量表达。

\paragraph{使役结构}  
句法学 \& 语义学 — 表示“使/让某人做…”的致使结构。

\paragraph{条件句}  
句法学 \& 语义学 — 表达“如果…，就…”条件关系的句子。

\paragraph{兼语句}  
句法学 — 一个名词在结构中既作宾语又作主语。

\paragraph{情态}  
语义学 \& 语用学 — 表示能力、必要性等的情态范畴。

\paragraph{时体标记}  
形态学 \& 语义学 — 标记时态或体的形式。

\paragraph{假设}  
语义学 \& 语用学 — 表示假设情景的表达。

\paragraph{受事主语句}  
句法学 \& 语义学 — 主语为动作承事者的句子。

\paragraph{可能}  
语义学 \& 语用学 — 表示可能性或潜在性的表达。

\paragraph{因果}  
语义学 \& 语用学 — 表示因果关系的表达。

\paragraph{并列}  
句法学 \& 语义学 — 平等地并列元素的结构。

\paragraph{明喻}  
语义学 \& 语用学 — 用“像”等词显性标记的比喻。

\paragraph{暗喻}  
语义学 \& 语用学 — 无显性比较词的隐喻。

\paragraph{比较}  
语义学 — 表示相似或差异的语言表达。

\paragraph{致使}  
句法学 \& 语义学 — 表示结果状态的致使表达。

\paragraph{让步}  
语义学 \& 语用学 — 虽承认…但仍…的让步关系。

\paragraph{转折}  
语义学 \& 语用学 — 标记对比或转折的关系。

\paragraph{递进}  
语义学 \& 语用学 — 表示进一步增强信息的关系。

\paragraph{指示}  
语义学 \& 语用学 — 根据上下文指示实体的表达。

\paragraph{话轮转换}  
语用学 — 对话中管理轮到谁发言的结构。

\paragraph{委婉语}  
语用学 — 缓和直接性的委婉表达。

\paragraph{拟人}  
语义学 \& 语用学 — 将人类特征赋予非人实体的表达。

\paragraph{夸张}  
语义学 \& 语用学 — 为强调而故意夸大的表达。

\paragraph{话语标记}  
语用学 — 引导和组织话语流程的词语。

\paragraph{礼貌}  
语用学 — 表示礼貌或维护面子策略的语言手段。

\paragraph{数量词}  
句法学 \& 语义学 — 数词加量词短语，表示确切数量。
\end{CJK*}
\section{Implementation Details}

We used 8 A100 GPUs with 80GB of memory for the experiments. While the exact GPU hours for each experiment were not precisely recorded, the total GPU usage did not exceed one hour. The system was set up with CUDA 12.4, Triton 3.0.0, and Ubuntu 22.04. For the Llama model, we employed the Hugging Face implementation of transformers, and for SAE model, we used the OpenSAE implementation\footnote{\url{https://github.com/THU-KEG/OpenSAE}} and set the hyperparameter $k$ to $128$ for TopK activation.

\end{document}